\documentclass{bmvc2k}
\usepackage{graphicx}
\usepackage{amsmath,amssymb} 
\usepackage{multirow}
\usepackage{booktabs}
\pdfoutput=1

\title{Few-shot Semantic Segmentation with Support-induced Graph Convolutional Network}

\addauthor{Jie Liu}{j.liu5@uva.nl}{1}
\addauthor{Yanqi Bao}{yanqibao1997@gmail.com}{2}
\addauthor{Wenzhe Yin}{w.yin@uva.nl}{1}
\addauthor{Haochen Wang}{h.wang3@uva.nl}{1}
\addauthor{Yang Gao}{gaoy@nju.edu.cn}{2}
\addauthor{Jan-Jakob Sonke}{j.sonke@nki.nl}{3}
\addauthor{Efstratios Gavves}{egavves@uva.nl}{1}

\addinstitution{
University of Amsterdam\\
Amsterdam, Netherlands
}
\addinstitution{
 Nanjing University\\ 
 Nanjing, China
}
\addinstitution{
The Netherlands Cancer Institute\\
Amsterdam, Netherlands
}
\runninghead{J. Liu, etal}{Few-shot semantic segmentation with
Support-induced GCN}

\begin{document}

\maketitle

\begin{abstract}
Few-shot semantic segmentation (FSS) aims to achieve novel objects segmentation with only a few annotated samples and has made great progress recently. 
Most of the existing FSS models focus on the feature matching between support and query to tackle FSS. 
However, the appearance variations between objects from the same category could be extremely large, leading to unreliable feature matching and query mask prediction.
To this end, we propose a Support-induced Graph Convolutional Network (SiGCN) to explicitly excavate latent context structure in query images.
%
%
%
Specifically, we propose a \emph{Support-induced Graph Reasoning (SiGR)} module to capture salient query object parts at different semantic levels with a Support-induced GCN. 
Furthermore, an \emph{instance association (IA)} module is designed to capture high-order instance context from both support and query instances.
By integrating the proposed two modules, SiGCN can learn rich query context representation, and thus being more robust to appearance variations. 
Extensive experiments on PASCAL-$5^i$ and COCO-$20^i$ demonstrate that our SiGCN achieves state-of-the-art performance.
\end{abstract}
\section{Introduction}
The development of deep convolutional neural networks has achieved great success in many computer vision tasks, such as image classification \cite{he2016deep, simonyan2014very}, semantic segmentation \cite{zhao2017pyramid, yu2015multi, dai2017deformable, zhang2021k}, and object detection \cite{han2021query, girshick2015fast, he2017mask, redmon2018yolov3}. Despite the effectiveness of deep networks, their over-reliance on large-scale annotated dataset still remains a fundamental limitation as data annotation requires large amount of human efforts,  especially for dense prediction tasks, \textit{e.g.,} semantic segmentation. To cope with these challenges, some weakly-supervised and semi-supervised semantic segmentation techniques \cite{shimoda2019self, lee2021bbam, chen2021semi} try to introduce weak annotations (e.g, points, scribbles, and bounding box) and unlabelled samples. However, these techniques usually fail to work when novel classes that never appear in the training set emerge. In such case, few-shot semantic segmentation (FSS) methods \cite{wu2021learning, nguyen2019feature, rakelly2018conditional, siam2019amp, boudiaf2021few, zhang2021self, yang2021mining, zhang2021few,liu2020part, Li_2021_CVPR, xie2021scale, min2021hypercorrelation, tian2020prior, Zhang_2019_CVPR,wang2019panet} struggle to segment novel objects with only a few pixel-wise annotated samples.

\begin{figure*}[!h]
	\centering
	\includegraphics[width=\linewidth]{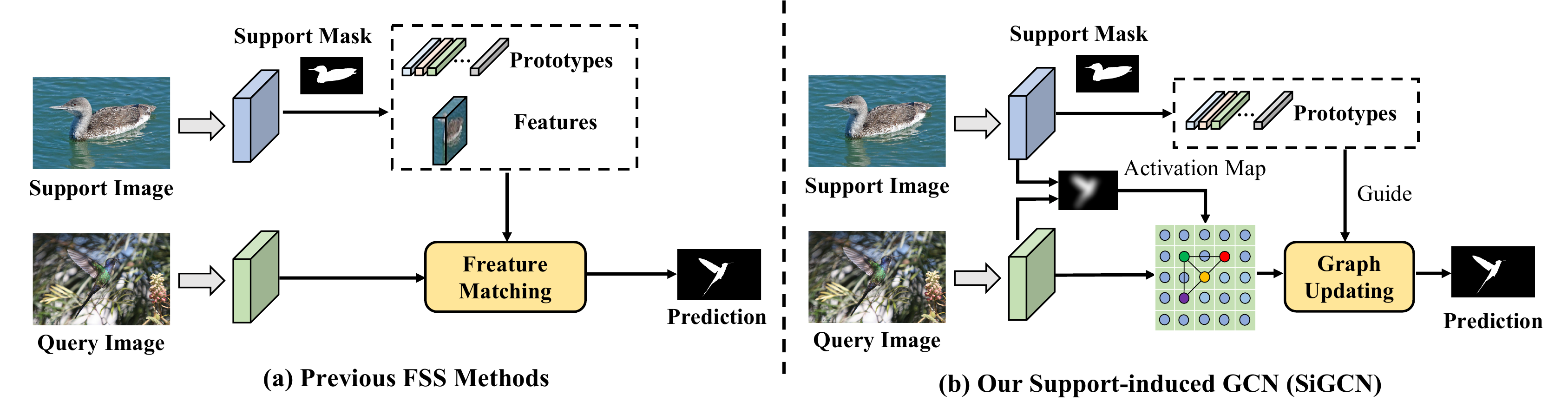}
	\caption{Comparison between (a) previous typical few-shot segmentation (FSS) methods and (b) the proposed support-induced graph convolutional network (SiGCN). (a) Previous FSS methods focus on designing effective feature matching mechanism between support prototypes/features and query feature. However, such matching techniques usually fail to work when the support and query objects exhibit large appearance variations. (b) Our SiGCN aims to address the appearance variation problem by explicitly mining query context information. Based on the activation map, SiGCN constructs a query graph to associate salient query object parts. Then the support prototypes are introduced to guide the graph updating process for effective query context excavation.}
	\label{fig1}
\end{figure*}

Meta-learning \cite{shaban2017one} is widely adopted in FSS to ensure better model generalization to novel classes. Specifically, both training and testing sets are composed of many sub-tasks (episodes), and each specific sub-task is divided into a support set and a query set, which share the same object classes. The FSS models aim to segment images from the query set given a few pixel-wise annotated images from the support set. Thus, the key challenge of the FSS task is how to leverage the information from the support set and the query set effectively, especially when the support and query objects exhibit large variations (e.g., appearance, scale, and location).

Although existing FSS methods \cite{tian2020prior, min2021hypercorrelation, zhang2021few} have achieved extensive progress, we observe a significant drop in performance when the support and query images share large appearance variance. For instance,
as shown in Figure \ref{fig1}, the birds in the support and query images have completely different appearance, which makes it difficult to segment the query image. Existing methods (Figure \ref{fig1} (a)) usually generate support foreground features or prototypes and then focus on designing various feature matching mechanisms between the support foreground features/prototypes and the query feature to achieve the query mask prediction. However, such matching with information coming only from the support set is suboptimal when it comes to predicting an accurate mask for the query image. By contrast, different query object parts (\textit{e.g.}, the wings and the main body of the bird) contain rich contextual information, which is beneficial to the query object segmentation. 

In this work, we propose a support-induced graph convolutional network (SiGCN) to tackle the aforementioned challenge. Our main idea is to explicitly enrich the query context representations with the guidance of support information in two ways. Firstly, we propose a novel support-induced graph reasoning (SiGR) module to extract query context from different query object parts. Specifically, as shown in Figure \ref{fig1}(b), we utilize the activation map, which is generated by the feature matching between the support and the query features, to capture local object parts that are discriminative and easily localized in the query image. These salient parts are associated together in the query graph by modeling the interaction between them, thus learning rich context in the query image. The query graph is then updated by the graph convolutional network, in which the state updating matrix is generated by the support prototypes to effectively extract query context. Secondly, an instance association (IA) module is introduced to aggregate query and support context from instance level. Specifically, given activation maps produced from different semantic levels, we  get corresponding updated query graphs (i.e, query features) in the SiGR module. These updated query features, which can be seen as different object instances, contain complementary context information. Therefore, we associate the query instances and the support instance together to learn high-order instance context. By proposing the SiGR and IA module, our SiGCN aggregates rich complementary context from both support and query images, thus being more robust to appearance variations.

In summary, our contributions are summarized as follows:
\begin{itemize}
\item We propose a \emph{support-induced graph convolution network (SiGCN)}, which utilizes the proposed IA and SiGR modules to capture complementary context from the query and the support set, for addressing the appearance variation problem in the FSS task. 

\item We propose a novel graph updating mechanism, in which support prototypes guide the node updating in the query graphs. This mechanism brings a significant performance gain to the proposed SiGCN.

\item Our method achieves state-of-the-art results on two FSS benchmarks, i.e., PASCAL-$5^i$ and COCO-$20^i$. Our SiGCN tackles the appearance variation problem from the perspective of extracting query contextual information, which sheds light for future research in this field. 

\end{itemize}
\section{Related Work}
\subsection{Semantic Segmentation} 
Semantic Segmentation is a fundamental computer vision task which aims to assign each pixel with a class label. Fully convolutional network (FCN) \cite{long2015fully} adopts fully convolutional layers for dense prediction and achieves superior performance. Inspired by FCN, many networks have been invented to boost the performance of semantic segmentation. Unet \cite{ronneberger2015u} presents an encoder-decoder structure to capture multi-scale context information. PSPNet \cite{zhao2017pyramid} utilizes  the pyramid pooling to aggregate more object details. Dilated convolution \cite{yu2015multi} and deformable convolution \cite{dai2017deformable} are introduced to FCN to further improve the semantic segmentation performance. Recently,  K-Net \cite{zhang2021k} proposes to segment both instance and semantic categories by a group of learnable kernels.  However, these methods still require tons of samples with pixel-wise annotations in the training stage. In our work, we focus on few-shot semantic segmentation.
\subsection{Few-shot Semantic Segmentation}
Few-shot semantic segmentation is proposed to segment novel objects given only few annotated samples. Most existing methods adopt episode-base meta-learning \cite{xie2021scale} strategy to ensure the efficient generalization of the model to the novel classes with few examples. Shaban et al. \cite{shaban2017one} proposes the first few-shot semantic segmentation work, which generates parameters from the conditioning branch and then segments the query image in the segmentation branch. Later on, prototype-based methods are proposed including CANet \cite{Zhang_2019_CVPR}, PANet \cite{wang2019panet}, SG-One \cite{zhang2020sg}, PPNet \cite{liu2020part}, PMMs \cite{yang2020prototype}. The intuition behind these methods lies in extracting representative foreground or background prototypes for support samples and then interacting these prototypes with query features using different strategies. Besides, matching-based methods \cite{tian2020prior, zhang2019pyramid, xie2021scale,min2021hypercorrelation, zhang2021few} also achieve great success in the few shot-segmentation task. PFENet \cite{tian2020prior} introduces a prior mask by implementing pixel-to-pixel matching between high-level support and query features, while PGNet \cite{zhang2019pyramid} and SAGNN \cite{xie2021scale} propose to implement element-wise matching between support and query by graph neural networks. Most recently, HSNet \cite{min2021hypercorrelation} adopts multi-scale dense matching to construct hypercorrelations and uses 4D convolutions to capture contextual information.  CyCTR \cite{zhang2021few} designs a cycle consistency transformer to aggregate pixel-wise support feature to query ones and achieves great success.  Different from previous work, we not only consider the matching between support and query, but also highlight context excavation in the query feature, our model is thus more robust to the object appearance variations.
\subsection{Graph Convolutional Network}
Graph convolutional networks (GCNs) \cite{kipf2016semi}, first proposed to tackle the semi-supervised classification, have seen massive application in computer vision tasks, e.g., action localization \cite{nawhal2021activity, zeng2019graph}, object detection \cite{meyer2021graph}, and semantic segmentation \cite{zhang2019dual}. Recently, GCNs are introduced to few-shot learning and achieve superior performance. Zhang et al \cite{zhang2021adargcn} propose an adaptive aggregation GCN (AdarGCN) to model the information passing between noisy samples and underlying training samples. Yan et al \cite{yan2022budget} deign a GCN based data selection policy to compute a context-sensitive representation for each unlabeled data by graph message passing. Furthermore,  han et al \cite{han2021query} introduce a heterogeneous GCN with three different types of edges to promote the pairwise matching for the few-shot object detection task. Inspired by the powerful relation modelling capacity of GCNs, we design a support-induced graph convolutional network (SiGCN) to enrich query contextual representation for addressing the object appearance variation problem in FSS.
\section{Method}
\begin{figure*}[!t]
	\centering
	\includegraphics[width=\linewidth]{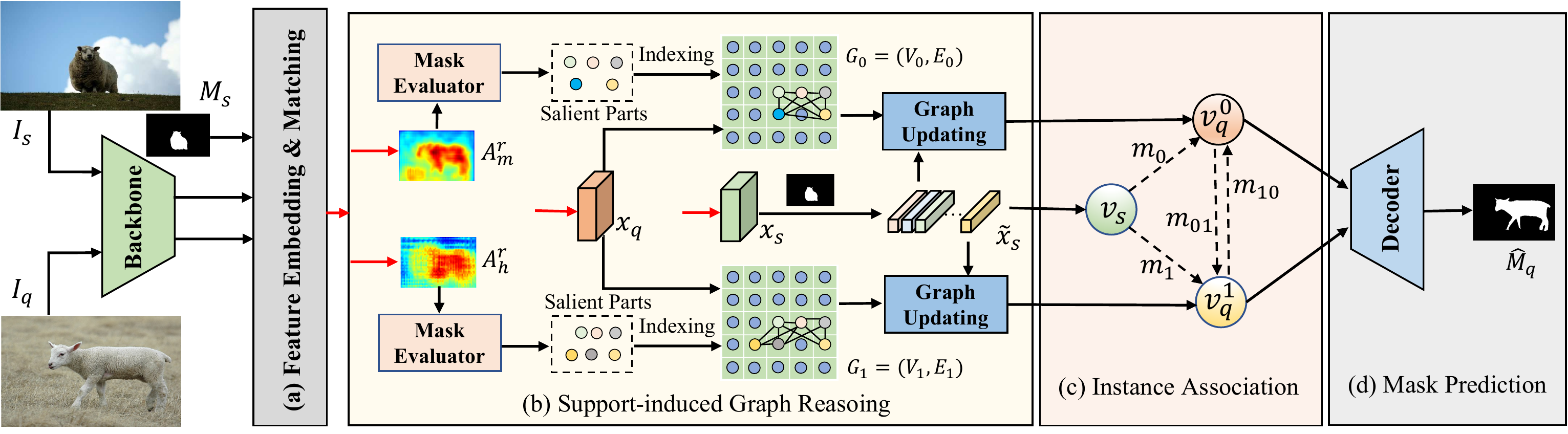}
	\caption{\textbf{Overall architecture of our proposed network.} Firstly, the feature embedding and matching network are introduced to generate support feature $x_s$, query feature $x_q$, middle-level activation map $A_m^r$, and high-level activation map $A_h^r$. Then, the support-induced graph reasoning module captures salient query object parts and associates them by a graph convolutoinal network. The support prototypes $\Tilde{x}_s$ is introduced to the graph updating for effective query context excavation. In addition, the instance association module aims to mine instance-level context by the massage passing between query instances (i.e., $v_q^0$ and $v_q^1$) and a support instance $v_s$. Finally, the updated features are concatenated and fed into a decoder for the final query mask prediction.}
	\label{fig2}
	\vspace{-4mm}
\end{figure*}
\subsection{Problem Setting}
To improve the model generalizability over previously unseen classes,  most methods adopt the episode-based meta-learning paradigm \cite{shaban2017one,vinyals2016matching}. 
Models are trained on the base classes $\mathcal{C}_{base}$ and tested on novel classes $\mathcal{C}_{novel}$. $\mathcal{C}_{base}$ and $\mathcal{C}_{novel}$ are disjoint, i.e., $\mathcal{C}_{base}\cap {C}_{novel}=\emptyset$.
Both training and inference are based on the episode data $(\mathcal{S},\mathcal{Q})$, which is composed of a support set $\mathcal{S}$, and a query set $\mathcal{Q}$ for a specific class $c$.
For the 1-way $K$-shot setting, the support set $\mathcal{S}=\{(I_s^{i}, M_s^{i})\}_{i=1}^K$ contains $K$ image-mask pairs, where $I_s^{i}$ represents $i$th support image and $M_s^{i}$ indicates corresponding binary mask. Similarly, we define the query set as $\mathcal{Q}=\{(I_q, M_q)\}$, where $I_q$ is query image and its binary mask $M_q$ is only available in the model training phase.
In the meta-training stage, the FSS model  takes input $\mathcal{S}$ and $I_q$ from a specific class $c$ and generates a predicted mask $\hat{M}_q$ for the query image.
Then, the model can be trained with the supervision of a binary cross-entropy loss between $M_q$ and $\hat{M}_q$, i.e., $\mathcal{L}_{BCE}(M_q, \hat{M}_q)$.
Finally, multiple episodes ${(S_i^{ts}, Q_i^{ts})}_{i=1}^{N_{ts}}$ are randomly sampled from the testing set $\mathcal{D}_{ts}$ for evaluation.
\subsection{Overview}
\label{sec3.3}
Figure \ref{fig2} illustrates the overall architecture of our proposed network, which consists of three main components: feature embedding and matching module, support-induced graph reasoning (SiGR) module and instance association (IA) module. Given the support set $\mathcal{S}=\{(I_s^{i}, M_s^{i})\}_{i=1}^K$ and query image $I_q$, we first encode them into feature maps by a weight-sharing backbone. Then, the feature embedding and matching network is introduced to generate support/query features $x_s\in\mathbb{R}^{C\times H \times W}$ and $x_q \in\mathbb{R}^{C\times H \times W}$, middle-level/high-level activation map $A_m^r\in\mathbb{R}^{H \times W}$ and $A_h^r \in\mathbb{R}^{H \times W}$, where $C$ and $HW$ are the channel and spatial dimension, respectively. We obtain activation maps by region-to-region matching between support and query features \cite{luo2021pfenet++}. Afterwards, the support-induced graph reasoning module is designed to enrich the query contextual representation by a novel support-induced graph convolutional network. Finally, we aggregate complementary context from both support and query by instance-level message passing in the instance association module.
\subsection{Support-induced Graph Reasoning}
The proposed SiGR module consists of two parallel branches. Taking the upper branch as shown in Figure \ref{fig2} as an example, we explain the support-induced graph reasoning process in details.

\noindent \textbf{Mask Evaluation.}
The coarse activation map for the query feature is usually not accurate enough and contains noise and non-salient parts. Therefore, we design a mask evaluator to select salient objects parts. Given the activation map $A_m^r$, we first process it with a min-max normalization to normalize the value to between 0 and 1.
Then, we select salient query object parts by introducing a salience matrix $S_m^r \in \mathbb{R}^{H\times W}$:
\begin{equation}
    S_m^r(i,j) = \left\{
            \begin{array}{rcl}
            1,     &      & {A_m^r(i, j) \geq t}\\
            0,     &      & {A_m^r(i, j) <t}
            \end{array}
            \right.,
\label{eq1}
\end{equation}
where $S_m^r$ is a sparse matrix, in which the elements with value 1 represent target location, otherwise noise or undiscriminating parts location. $t$ is a threshold which decides whether a pixel belongs to salient parts or not.

\noindent \textbf{Graph Construction.}
Based on the defined salience matrix $S_m^r$, we select $N_m=\sum_{i,j}^{HW}{S_m^r(i, j)}$ most salient pixels in the query feature $x_q$ as the set of salient parts $P_m=\{p_m^n\}_{n=1}^{N_m}$, which are discriminative local parts of the object for segmentation. The captured salient object parts are further associated together by a fully connected graph to learn rich query contextual information.
Specifically, we treat each pixel in the query feature as a node and build the graph $G_0 = (V_0, E_0)$, where $V_0$ and $E_0$ are the node set the edge set, respectively. For the connecting edges, We adopt fully-connected edges between $N_m$ salient parts specified in $P_m$. To achieve better generalization over novel objects, the edge weights are designed to be the similarity score between two connected nodes. Thus, the adjacency matrix $A_0\in \mathbb{R}^{N \times N}$, $N=HW$, for the graph $G_0$ is constructed as:
\begin{equation}
    A_0(i,j) = \left\{
            \begin{array}{rcl}
            \frac{v_i^\mathrm{T}v_j}{\parallel v_i \parallel  \parallel v_j \parallel},     &      & {\text{if edge}<i, j>\in \mathcal{E}}\\
            0,   &      & {\text{else}}
            \end{array}
            \right.,
\label{eq2}
\end{equation}
where $v_i\in\mathbb{R}^C$ and $v_j\in\mathbb{R}^C$ are two node vectors in the connection, $\mathcal{E}$ is the edge connection set of all the salient parts feature vectors. 
%

%
\noindent \textbf{Graph Updating.}
To update graphs by aggregating the salient parts' information, we propose a support-induced graph convolution network (SiGCN) to achieve effective information aggregation with the guidance of support foreground prototypes.
For the query graph $V_0$, we first construct a two-layer GCN to perform the node state updating. Formally, the $l$-th layer graph convolution is formulated as:
\begin{equation}
X^{l+1} = \sigma^l(\hat{A}X^{l}W^l),
\label{eq3}
\end{equation}
where $X^l\in \mathbb{R}^{N\times d_l}$ and  $X^{l+1}\in \mathbb{R}^{N\times d_{l+1}}$ are the input and output feature of all nodes at the $l$-th layer, respectively,  $d_l$ and $d_{l+1}$ are the corresponding feature dimensions, $d_0$ is equal to the node feature dimension C. $\hat{A}=\Tilde{D}^{-\frac{1}{2}}\Tilde{A}\Tilde{D}^{\frac{1}{2}}$ is the normalized adjacency matrix \cite{kipf2016semi}, where $\Tilde{A}=A+I$ and $\Tilde{D}$ is the diagonal degree matrix of $\Tilde{A}$. $W^l \in \mathbb{R}^{d_l\times d_{l+1}}$ and $\sigma^l(\cdot)$ denote the state updating matrix and activation function at the layer $l$, respectively.

Normally, the state updating matrix $W$ is optimized with the model training proceeds. Chen et. al \cite{chen2019graph} propose to adopt a 1D convolution as state updating function. Inspired by this, we implement the state updating matrix as a 1D convolution with fixed kernel generated from the support foreground vectors.

\begin{equation}
X^{l+1} = \sigma^l(\hat{A}\mathcal{F}_c^l(X^{l}|\theta^l)),
\label{eq4}
\end{equation}
where $\mathcal{F}_c^l(\cdot)$ is the 1D convolution with fixed kernel parameters $\theta^l$ at the $l$-th layer. $\theta \in \mathbb{R}^{k\times C}$, where $k$ is the number of support prototypes (i.e., kernel size of 1D convolution), is generated by applying an average pooling to the support foreground feature.
Given the high-level activation map $A_h^r$, we can similarly select salient query object parts, construct the query graph $G_1=(V_1, E_1)$, and implement support-induced graph updating. Finally, we can obtain two updated query graphs, which aggregate rich query contextual representation with the guidance of support prototypes.

\subsection{Instance Association}
\label{IAM}
Although rich context has been aggregated into query graphs, the complementary inter-graph context still remains unexploited. Therefore, the updated query graphs are further transformed into query instance features $v_q^0 \in \mathbb{R}^{C\times H \times W}$ and $v_q^1\in \mathbb{R}^{C\times H \times W}$, respectively. Meanwhile, we can also adopt the 1D average pooling and the reshape operation over the support foreground prototypes $\Tilde{x}_s$ to get the support instance feature $v_s\in \mathbb{R}^{C\times s\times s }$, where $s$ is the spatial size. Then, the the query instance features and support instance feature are associated together by the instance association module to learn instance-level contextual information.
For the query instance feature $v_q^0$, the information gathered from the support instance feature $v_s$ and the query instance feature $v_q^1$ can be formulated as:
\begin{equation}
m_0 = \mathcal{R}(v_s)\mathcal{R}(v_s)^\mathrm{T}\mathcal{R}(v_q^0),
m_{10} = \mathcal{R}(v_q^1)\mathcal{R}(v_q^1)^\mathrm{T}\mathcal{R}(v_q^0),
\label{eq5}
\end{equation}
where $m_0\in\mathbb{R}^{C\times HW}$ and $m_{10}\in\mathbb{R}^{C\times HW}$ are the instance-level contextual information from support instance and query instance, respectively. $\mathcal{R}(\cdot)$ is the vectorization operation. Then, the query instance feature can be updated as $\Tilde{v}_q^0$ by:
\begin{equation}
\Tilde{v}_q^0 = \frac{1}{2}(v_q^0+(\alpha m_0 + \beta m_{10})),
\label{eq6}
\end{equation}
where $\alpha$ and $\beta$ are two hyperparameters to weight the contribution of the contextual information from the support and query, respectively. Similarly, the other query instance feature $v_q^1$ can be updated as $\Tilde{v}_q^1$. Compared with previous state, $\Tilde{v}_q^0$ and $\Tilde{v}_q^1$ aggregate more instance-level context information from both support and query, thus being more robust to the appearance variations.
\subsection{ Mask Prediction and Model Training}
Finally, we feed the updated query instance features as well as activation maps specified in Section \ref{sec3.3} to make pixel-wise prediction for query.
\begin{equation}
\hat{M}_q = \mathcal{F}_{decoder}(\mathcal{C}(\Tilde{v}_q^0, \Tilde{v}_q^1, A_m^p, A_m^r, A_h^p, A_h^r)),
\label{eq7}
\end{equation}
where $\mathcal{C}(\cdot)$ is the channel-wise concatenation. $\mathcal{F}_{decoder}(\cdot)$ is the decoder network, which consists of a Atrous Spatial Pyramid Pooling (ASPP) module \cite{chen2017rethinking} and three consecutive convolution layers with residual connection. For the activation maps $A_m^p$ and $A_h^p$, the superscript indicates that the activation map comes from pixel-to-pixel matching (p) \cite{tian2020prior} between support and query features. To generate the final segmentation map, we upsample $\hat{M}_q$ by bilinear interpolation. Finally, we adopt the binary cross entropy loss to train our model on all the episodes in the training set $\mathcal{D}_{tr}$.

\section{Experiments}
\subsection{Implementation Details}
Following previous methods \cite{tian2020prior,min2021hypercorrelation}, we conduct experiments on PASCAL-$5^i$ and COCO-$20^i$. Classes in each dataset are divided into four base-novel class folds for cross validation. Our model is based on PFENet \cite{tian2020prior} and we also adopt the same training and evaluation setting as PFENet. We train our model with SGD optimizer on PASCAL-$5^i$ for 200 epochs and COCO-$20^i$ for 50 epochs. The learning rate is initialized as 0.005 with batch size 8 and decays according to the  poly learning rate scheduler. During evaluation, we adopt the mIoU as main metric and also report FBIoU. For the K-shot setting, we fuse foreground prototypes from each support image-mask pair to generate $\theta$ in Eq. (\ref{eq4}) and the support instance feature $v_s$ in the instance association module. 
\begin{table*}[!t]
\centering
\caption{Comparison with the state-of-the-art methods on PASCAL-$5^i$ dataset under both 1-shot and 5-shot settings. The mIoU of each fold, and the averaged mIoU \& FB-IoU of all folds are reported.}
\renewcommand\arraystretch{1}
\resizebox{\linewidth}{!}{
\begin{tabular}{r|c|cccccc|cccccc}
\toprule[2pt]
\multirow{2}{*}{Methods} & \multicolumn{1}{c|}{\multirow{2}{*}{Backbone}} & \multicolumn{6}{c|}{1-shot}  & \multicolumn{6}{c}{5-shot}  \\
& \multicolumn{1}{c|}{} 
& \multicolumn{1}{l}{Fold-0} & Fold-1  & Fold-2  & Fold-3   & Mean  & FB-IoU
& \multicolumn{1}{l}{Fold-0} & Fold-1  & Fold-2  & Fold-3   & Mean  & FB-IoU \\ \hline
OSLSM (BMVC'17) \cite{shaban2017one} & VGG16                         
& 33.6  & 55.3  & 40.9  & 33.5   & 40.8  & 61.3                 
& 35.9  & 58.1  & 42.7  & 39.1   & 43.9  & 61.5           \\
co-FCN (ICLRW'18) \cite{rakelly2018conditional}  & VGG16
& 36.7  & 50.6  & 44.9  & 32.4   & 41.1  & 60.1   
& 37.5  & 50.0  & 44.1  & 33.9   & 41.4  & 60.2            \\
PFENet (TPAMI’20) \cite{tian2020prior}  &  VGG16
& 56.9  & 68.2  & 54.4  & 52.4   & 58.0  & 72.0  
& 59.0  & 69.1  & 54.8  & 52.9   & 59.0  & 72.3            \\
HSNet (ICCV'21) \cite{min2021hypercorrelation}  & VGG16
& 59.6  & 65.7  & 59.6  & 54.0   & 59.7  & 73.4 
& 64.9  & 69.0  & 64.1  & 58.6   & 64.1  & 76.6            \\ \hline
PFENet (TPAMI’20) \cite{tian2020prior}  & ResNet50
& 61.7  & 69.5  & 55.4  & 56.3   & 60.8  & 73.3  
& 63.1  & 70.7  & 55.8  & 57.9   & 61.9  & 73.9            \\
RePRI (CVPR'21) \cite{boudiaf2021few}   &  ResNet50
& 59.8  & 68.3  & 62.1  & 48.5   & 59.7  & -    
& 64.6  & 71.4  & 71.1  & 59.3   & 66.6  & -               \\
SAGNN (CVPR'21) \cite{xie2021scale}  & ResNet50
& 64.7  & 69.6  & 57.0  & 57.3   & 62.1  & 73.2 
& 64.9  & 70.0  & 57.0  & 59.3   & 62.8  & 73.3            \\
MMNet (ICCV'21) \cite{wu2021learning}  & ResNet50
& 62.7  & 70.2  & 57.3  & 57.0   & 61.8  & -
& 62.2  & 71.5  & 57.5  & 62.4   & 63.4  & -               \\
HSNet (ICCV'21) \cite{min2021hypercorrelation}  & ResNet50
& 64.3  & 70.7  & 60.3  & 60.5   & 64.0  & 76.7  
& 70.3  & 73.2  & 67.4  & 67.1   & 69.5  & 80.6             \\
CyCTR (NIPS'21) \cite{zhang2021few}  & ResNet50
& 67.8  & 72.8  & 58.0  & 58.0   & 64.2  & -  
& 71.1  & 73.2  & 60.5 & 57.5   & 65.6  & -             \\ \hline
\textbf{Baseline}    & ResNet50
& 62.5  & 69.4  & 58.9   & 55.9   & 61.7   & 72.0  
& 63.2  & 70.5  & 60.1   & 57.6   & 62.9   & 74.4                                      \\
\textbf{SiGCN} & ResNet50
& 65.1  & 70.1   & 65.2  & 60.8  & \textbf{65.3} & \textbf{77.5}
& 68.9  & 72.6   & 66.8  & 65.8  & \textbf{68.5}  & \textbf{78.3 }             \\ 
\bottomrule[2pt]
\end{tabular}}
\vspace{-3mm}
\label{pascal}
\end{table*}
\begin{table*}[!t]
\centering
\caption{Comparison with the state-of-the-art methods on COCO-$20^i$ dataset under both 1-shot and 5-shot settings. The mIoU of each fold, and the averaged mIoU \& FB-IoU of all folds are reported.}
\renewcommand\arraystretch{1}
\resizebox{\linewidth}{!}{
\begin{tabular}{r|c|cccccc|cccccc}
\toprule[2pt]
\multirow{2}{*}{Methods} & \multicolumn{1}{c|}{\multirow{2}{*}{Backbone}} & \multicolumn{6}{c|}{1-shot}  & \multicolumn{6}{c}{5-shot}  \\
& \multicolumn{1}{c|}{} 
& \multicolumn{1}{l}{Fold-0} & Fold-1  & Fold-2  & Fold-3   & Mean  & FB-IoU
& \multicolumn{1}{l}{Fold-0} & Fold-1  & Fold-2  & Fold-3   & Mean  & FB-IoU \\ \hline
FWB(ICCV'19) \cite{nguyen2019feature} & VGG16    
& 18.4   & 16.7   & 19.6   & 25.4   & 20.0 & -     
& 20.9   & 19.2   & 21.9   & 28.4   & 22.6 & -    \\
PFENet(TPAMI’20) \cite{tian2020prior} & VGG16
& 33.4  & 36.0    & 34.1   & 32.8   & 34.1 & 60.0   
& 35.9  & 40.7   & 38.1   & 36.1   & 37.7 & 61.6    \\
SAGNN(CVPR'21) \cite{xie2021scale} & VGG16  
& 35.0  & 40.5   & 37.6   & 36.0   & 37.3 & 61.2  
& 37.2  & 45.2   & 40.4   & 40.0   & 40.7 & 63.1  \\ 
\hline
RePRI(CVPR'21) \cite{boudiaf2021few} & ResNet50  
& 31.2  & 38.1   & 33.3   & 33.0   & 34.0 & -      
& 38.5  & 46.2   & 40.0   & 43.6   & 42.1 & -    \\
SAGNN(CVPR'21) \cite{xie2021scale} & ResNet101 
& 36.1  & 41.0   & 38.2   & 33.5   & 37.2 & 60.9  
& 40.9  & 48.3   & 42.6   & 38.9   & 42.7 & 63.4    \\ 
MMNet(ICCV'21) \cite{wu2021learning}  & ResNet50 
& 34.9  & 41.0   & 37.2   & 37.0   & 37.5 & -  
& 37.0  & 40.3   & 39.3   & 36.0   & 38.2 & -     \\
HSNet(ICCV'21) \cite{min2021hypercorrelation} & ResNet50  
& 36.3  & 43.1   & 38.7   & 38.7   & 39.2 & 68.2  
& 43.3  & 51.3   & 48.2   & 45.0   & 46.9 & 70.7  \\
CyCTR(NIPS'21) \cite{zhang2021few} & ResNet50  
& 38.9  & 43.0   & 39.6   & 39.8   & 40.3 & -  
& 41.1  & 48.9   & 45.2   & 47.0   & 45.6 & -   \\ \hline
\textbf{Baseline} & ResNet50                    
& 33.6   & 39.2    & 36.5    & 34.2   & 35.9    & 60.7 
& 35.3   & 43.1    & 38.4    & 38.7   & 38.9    & 62.0        \\
\textbf{SiGCN} & ResNet50
& 38.7  & 46.3  & 43.1  & 37.5 & \textbf{41.4}  & \textbf{62.7}
& 44.9  & 54.5  & 46.5  & 45.9 & \textbf{48.0}  & \textbf{66.2} \\ 
\bottomrule[2pt]
\end{tabular}}
\vspace{-4mm}
\label{coco}
\end{table*}
\subsection{Comparison with the State-of-the-art methods}
\begin{figure*}[!t]
	\centering
	\includegraphics[width=\textwidth]{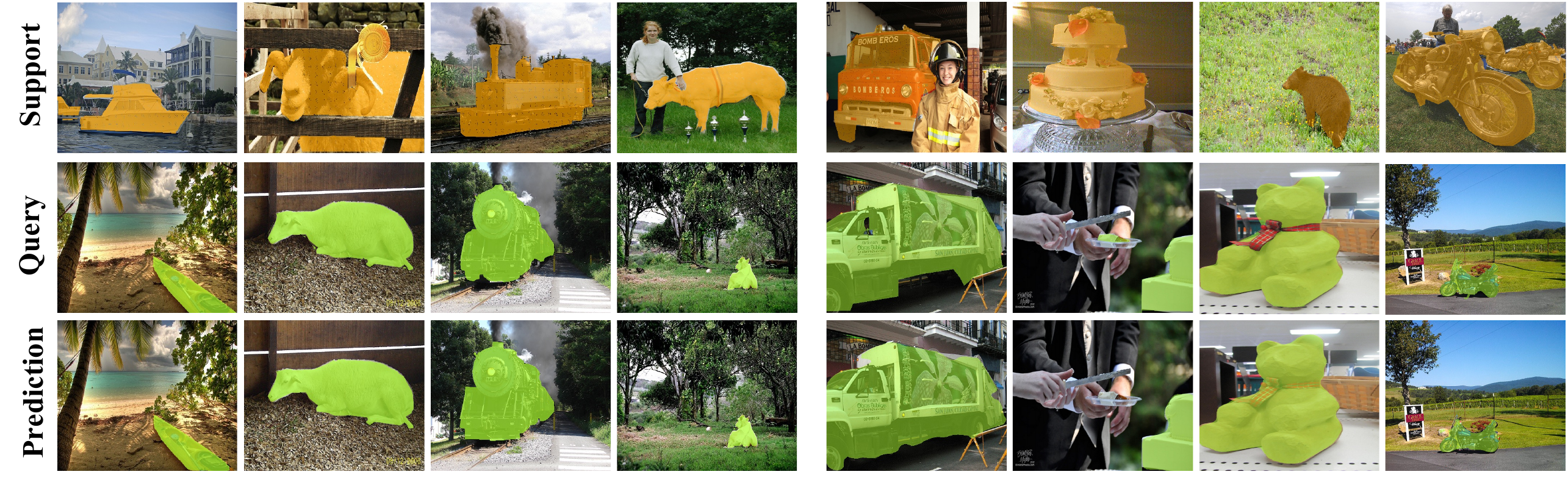}
	\caption{Qualitative results on the PASCAL-$5^i$ (left four columns) and COCO-$20^i$ (right four columns) under the 1-shot setting. Our SiGCN can make accurate segmentation prediction, even though the support and query objects exhibit larger appearance variations.\vspace{-4mm}}
	\label{fig3}
\end{figure*}

\textbf{PASCAL-$5^i$.}
In Table \ref{pascal}, we show the performance comparison between the proposed SiGCN and the leading methods on the PASCAL-$5^i$ dataset. With the ResNet50 as backbone, we can conclude that the proposed SiGCN achieves the state-of-the-art performance under the 1-shot setting and competitive result under the 5-shot setting. Specifically, SiGCN surpasses previous SOTA (i.e., CyCTR) by 1.1\% in the 1-shot setting, while achieving competitive results as previous SOTA HSNet (68.5 \emph{vs} 69.5) in the 5-shot setting. Additionally, compared with the baseline model, which is constructed by removing the SiGR and IA modules from SiGCN (see complementary materials for more details), our SiGCN achieves significant improvement, i.e., 3.6\% and 6.4\% under the 1-shot and 5-shot settings, respectively. Considering the comparison with previous SOTA methods using the FB-IoU, SiGCN also obtain competitive results (77.5 and 78.3 under the 1-shot and 5-shot setting, respectively).
\textbf{COCO-$20^i$.}
Table \ref{coco} shows the comparison between previous SOTA methods and our SiGCN on the COCO-$20^i$ dataset. Although the COCO-$20^i$ is more challenging (larger appearance and scale variations than PASCAL-$5^i$), SiGCN also achieves state-of-the art results (41.4 and 48.0 under 1-shot and 5-shot settings, respectively). It demonstrates the superiority of our proposed SiGCN in addressing the appearance variation problem. Furthermore, SiGCN with ResNet50 backbone achieves significant mIoU improvement (i.e., 5.5\% and 9.1\% under the 1-shot and 5-shot settings, respectively) over the baseline model.\\
\textbf{Qualitative Results.} In Figure \ref{fig3}, we report some typical qualitative results generated by our SiGCN. As can be seen, the support and query object exhibits large appearance variations, but our  approach can also produce accurate mask prediction for the query image, which thanks to the query context mining by our method. More experimental and quantitative results can be found in the supplementary material.\vspace{2mm}

\begin{minipage}[h]{\textwidth}
\begin{minipage}{0.45\textwidth}
\makeatletter\def\@captype{table}
\begin{center}
\caption{Effect of key modules in SiGCN.}\vspace{1mm}
\centering
\resizebox{5.5cm}{!}{
\begin{tabular}{cc|ccccc|c}
\toprule[2pt]
\multirow{2}{*}{SiGR} & \multirow{2}{*}{IA} & \multicolumn{5}{c|}{1-shot mIoU} \\ 
                     &                      & $5^0$    & $5^1$    & $5^2$    & $5^3$    & mean & \multirow{-2}{*}{FB-IoU} \\ \hline
                     &                      & 62.5 & 69.4 & 58.9 & 55.9 & 61.7 & 72.0    \\
                     & \checkmark           & 62.0 & 70.0 & 58.5 & 59.4 & 62.7 & 73.5   \\
\checkmark           &                      & 64.6 & 69.8 & 64.1 & 60.8 & 64.8 & 77.4   \\
\checkmark           & \checkmark           & 65.1 & 70.1 & 65.2 & 60.8 & \textbf{65.3} & \textbf{77.5}   \\ \bottomrule[2pt]
\end{tabular}\vspace{-2mm}}
\label{components}
\end{center}
\end{minipage}\hspace{3mm}
\begin{minipage}{0.45\textwidth}
\makeatletter\def\@captype{table}
\begin{center}
\caption{Effect of different GCN variants.}\vspace{-3mm}
\resizebox{6.1cm}{!}{
\begin{tabular}{c|ccccc|c}
\toprule[2pt]
& \multicolumn{5}{c|}{1-shot mIoU} & \\ 
\multirow{-2}{*}{GCN Variants} & $5^0$ & $5^1$ & $5^2$ & $5^3$ & mean    
& \multirow{-2}{*}{FB-IoU} \\ \hline
baseline      & 62.0   & 70.0   & 58.5  & 59.4  & 62.7   & 73.5  \\
GCN \cite{kipf2016semi}      & 63.0   & 69.4   & 59.7  & 57.1  & 62.3  & 73.3  \\
SiGCN-1      & 63.4   &  69.6  & 62.2  & 60.0  & 63.8   & 75.4 \\
SiGCN-2      & 65.1   & 70.1   & 65.2  & 60.8  & \textbf{65.3}  &  \textbf{77.5}\\
\bottomrule[2pt]
\end{tabular}}
\label{gcn}
\end{center}
\end{minipage}
\end{minipage}

\subsection{Ablation Study}

\textbf{Component analysis.}
There are two main components in our model, namely the support-induced graph reasoning (SiGR) module  and the instance association (IA) module . We validate the effectiveness of each component and present the results in Table  \ref{components}. It can be summarized that the SiGR module plays the most important role in the performance improvement while the IA module is indispensable. With SiGR and IA module, our SiGCN achieves state-of-the-art performance.

\textbf{Effect of different GCN variants.} To investigate the effect of different GCN variants in the target association module, we adopt three GCN variants, i.e., plain GCN, SiGCN with one and two layers (denoted as SiGCN-1 and SiGCN-2, respectively). As shown in Table \ref{gcn}, our model with plain GCN achieves sightly better performance than the baseline model, while our model with SiGCN makes substantial performance improvement as  the number of the SiGCN layer increasing from one to two.
\begin{figure*}[!t]
	\centering
	\includegraphics[width=0.7\textwidth,height=0.25\textwidth]{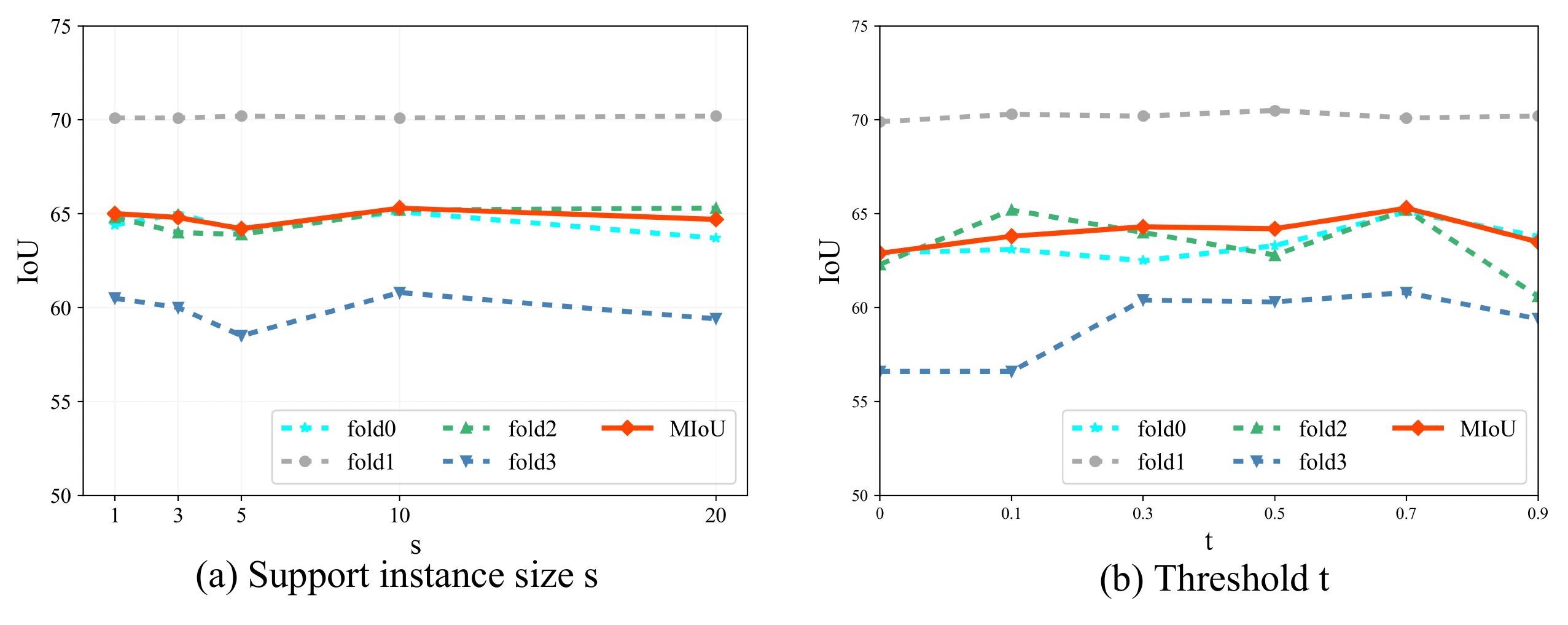}\vspace{-2mm}
	\caption{Ablation studies of (a) support instance size $s$ and (b) mask threshold $t$ on PASCAL-$5^i$ under the 1-shot setting.}\vspace{-4mm}
	\label{fig4}
\end{figure*}

\begin{minipage}[h]{\textwidth}
\begin{minipage}{0.45\textwidth}
\makeatletter\def\@captype{table}
\begin{center}
\centering
\caption{Effect of $k$ prototypes.}\vspace{1mm}
\resizebox{5.5cm}{!}{
\begin{tabular}{c|ccccc|c}
\toprule[2pt]
& \multicolumn{5}{c|}{1-shot mIoU} & \\ 
\multirow{-2}{*}{$k$ prototypes} & $5^0$ & $5^1$ & $5^2$ & $5^3$ & mean    
& \multirow{-2}{*}{FB-IoU} \\ \hline
1      & 64.8   & 69.7      & 63.5      & 59.0       &  64.3     & 75.9  \\
3      & 64.1   & 70.2   & 64.3  & 59.5  & 64.5  & 76.2  \\
5      & 65.1   & 70.1   & 65.2  & 60.8  & \textbf{65.3}  & \textbf{77.5}\\
7      & 63.7   & 69.8       & 63.8     & 60.2       &  64.4     & 75.5  \\
\bottomrule[2pt]
\end{tabular}}

\label{kernel}
\end{center}
\end{minipage}\hspace{3mm}
\begin{minipage}{0.45\textwidth}
\makeatletter\def\@captype{table}
\begin{center}
\centering
\caption{Effect of different backbones.}\vspace{-3mm}
\resizebox{6.2cm}{!}{
\begin{tabular}{c|ccccc|c}
\toprule[2pt]
& \multicolumn{5}{c|}{1-shot mIoU} & \\ 
\multirow{-2}{*}{Backbone} & $5^0$ & $5^1$ & $5^3$ & $5^3$ & mean    
& \multirow{-2}{*}{FB-IoU} \\ \hline
VGG16       & 58.0   & 68.0   & 62.9  & 54.2  & 60.8   & 73.5  \\
ResNet-50   & 65.1   & 70.1   & 65.2  & 60.8  & 65.3  &  77.5\\
ResNet-101   & 65.4   & 71.0   & 64.6  & 61.7  & \textbf{65.7}   & \textbf{78.3}  \\
\bottomrule[2pt]
\end{tabular}}
\label{backbone}
\end{center}
\end{minipage}
\end{minipage}

\textbf{Influence of $k$ support prototypes.}
The number of support prototypes $k$ in Eq. (\ref{eq4}) is a key parameter in SiGCN, which determines the quality of support foreground used for guiding the query graph updating. We vary $k$ in $\{1, 3, 5, 7\}$ to observe the performance of our model. In Table \ref{kernel}, $k=5$ achieves best performance, which indicates that SiGCN with 5 support foreground prototypes provides better guidance to query graphs for capturing contextual information. 

\textbf{Influence of support instance size $s$.}
The support instance feature size $s$ in the instance association module (section \ref{IAM}) determines the context information in the support instance. The value of $s$ is taken from $\{1, 3, 5, 10, 20\}$ to observe the performance variations. As in Figure \ref{fig4}(a), SiGCN with $s=10$ exhibits the best performance, thus we set $s=10$ in all of our experiments.

\textbf{Influence of mask threshold $t$.}
The mask threshold $t$ in Eq.(\ref{eq1}) is essential for selecting discriminative parts, i.e., targets in the query graph. Thus we take its value from {0,0.1, 0.3, 0.5, 0.7, 0.9} to investigate the performance of our SiGCN. As can be seen in Figure \ref{fig4}(b), SiGCN with $t=0.7$ performs best. Additionally, SiGCN without mask threshold in Eq.(\ref{eq5}) (i.e., $t=0$) also shows competitive results.

\textbf{Effect of different backbones.}
To evaluate the performance of our model with different backbones, we adopt three backbone networks (i.e., VGG16, ResNet50, and ResNet101) to implement experiments. As shown in Table \ref{backbone}, our SiGCN with ResNet101 as backbone yields superior performance.
\section{Conclusion}
In this work, we propose a support-induced graph convolutional network (SiGCN) to tackle the challenging appearance variation problem for the few-shot semantic segmentation (FSS) task. Our main idea is to enrich query context representation with the discriminative parts in the query objects and support objects. To achieve this, we propose a support-induced graph reasoning (SiGR) module to associate salient query object parts with the graph convolution network, in which support prototypes is introduced to guide the graph updating process. Additionally, an instance association (IA) module is designed to capture high-order context from the support instances and complementary query instances simultaneously. Extensive experiments on two FSS benchmarks prove that our SiGCN achieves state-of-the-art performance under both 1-shot and 5-shot settings. We believe that the idea of explicitly mining query context will shed light for future research in this field.

\paragraph{Acknowledgement} This work was partially funded by Elekta Oncology Systems AB and a RVO public-private partnership grant (PPS2102).

\bibliography{egbib}

\newpage
\section{Supplementary Materials}
In this supplementary material, we firstly present extensive implementation details about our experiments, and then give two more additional ablation studies and analysis about different level features and matching methods for activation maps. Finally, we provide more qualitative visualization results for PASCAL-$5^i$ and COCO-$20^i$ benchmarks under the large object appearance and scale variations scenarios.

\textbf{Implementation Details}
In our experiments, we employ ResNet-50 (VGG-16 and ResNet-101) pre-trained on ImageNet as our backbone network.
For ResNet-50 and ResNet-101, the dilation convolution is introduced to ensure that the feature receptive fields of layer2, layer3, and layer4 preserve the same spatial resolution. 
The backbone weights are frozen except for layer4, which is required to learn more robust activation maps. The proposed model is validated on PASCAL-$5^i$ and COCO-$20^i$ benchmarks, which are widely used in the few-shot semantic segmentation for cross validations. The detailed split of testing classes for each cross validation (fold) is shown in Table \ref{pascal_class} and Table \ref{coco_class}, respectively.

The model is trained with a SGD optimizer for 200 and 50 epochs on the PASCAL-$5^i$ and the COCO-$20^i$ benchmarks, respectively.
The learning rates are initialized as 0.005 and 0.002 with a poly learning rate schedule in PASCAL-$5^i$ and COCO-$20^i$, respectively. Our entire network is trained with the same learning rate during each epoch, while layer4 of the backbone network should be ensured a lower learning rate for fine-tuning, thus its parameters starts back-propagation after training for multiple epochs. And the batch size is set as 8 on PASCAL-$5^i$ and 32 on COCO-$20^i$.
Data augmentation strategies like random rotation and flip are adopted in the training stage, and all images are cropped to $473\times473$ patches for two benchmarks. Besides, no post-processing is used on PASCAL-$5^i$, while for COCO-$20^i$, we adopt the multi-scale testing strategy for the model evaluation due to extremely large object appearance and scale variations. By the way, the original Groudtruth of the query image without any resize operations is adopted for the model evaluation.

In addition, the support instance size $s$ in the IA module are set as 10, and mask threshold $t$ are set as 0.7. For the SiGR module, we set the number of support prototypes $k=5$ in each SiGCN layer. Finally, we implement our model with PyTorch 1.10.0 and conduct all the experiments with Nvidia Tesla A100 GPUs and CUDA11.3.

\begin{table}[!h]
\caption{Testing classes split for each fold in PASCAL-$5^i$ dataset.}
\begin{center}
\label{pascal_class}
\resizebox{8cm}{!}{
\begin{tabular}{cc}
\hline
Fold   & Testing (novel) classes \\ \hline
Fold-0 & Aeroplane, Bicycle, Bird, Boat, Bottle     \\
Fold-1 & Bus, Car, Cat, Chair, Cow                \\
Fold-2 & Diningtable, Dog, Horse, Motorbike, Person                \\
Fold-3 & Potted plant, Sheep, Sofa, Train, Tvmonitor                \\ \hline
\end{tabular}}
\end{center}
\end{table}
\newcommand{\tabincell}[2]{\begin{tabular}{@{}#1@{}}#2\end{tabular}}
\begin{table}[!h]
\caption{Testing classes split for each fold in COCO-$20^i$ dataset.}
\begin{center}
\label{coco_class}
\resizebox{\textwidth}{!}{
\begin{tabular}{cc}
\hline
Fold   & Testing (novel) classes \\ \hline
Fold-0 & \tabincell{c}{Person, Airplane, Boat, Parking meter, Dog, Elephant, Backpack,Suitcase, Sports Ball, \\ Skateboard, Wine glass, Spoon, Sandwich, Hot dog, Chair, Dining table, Mouse, Microwave, Scissorse} \\
Fold-1 & \tabincell{c}{Bicycle, Bus, Traffic light, Bench, Horse, Bear, Umbrella, Frisbee, Kite, Surfboard , \\ Cup, Bowl, Orange, Pizza, Couch, Toilet, Remote, Oven, Book, Teddy bear}\\
Fold-2 & \tabincell{c}{Car, Train, Fire hydrant, Bird, Sheep, Zebra, Handbag, Skis, Baseball bat, Tennis racket, \\ Fork, Banana, Broccoli, Donut, Potted plant, Tv, Keyboard, Sink, Toaster, Clock, Hair drier}                \\
Fold-3 & \tabincell{c}{Motorcycle, Truck, Stop sign, Cat, Cow, Giraffe, Tie, Snowboard, Baseball glove, Bottle, \\ Knife, Apple, Carrot, Cake, Bed, Laptop, Cell phone, Refrigerator, Vase, Toothbrush} \\ \hline
\end{tabular}}
\end{center}
\end{table}

\textbf{Effect of activation maps from different feature levels.}
To evaluate the quality of activation maps generated from different feature levels, we adopt three different-level features, i.e., block3 (middle-level), block4 (high-level), and block3\&4 (middle-level\& high-level). Support and query features from each level are matched by pixel-to-pixel matching to generate corresponding activation map. As shown in Table \ref{mask}, our model with the activation map generated from block3\&4 yields superior performance, while our model with the activation map generated from block3 or block4 also achieves competitive results.

\textbf{Effect of region matching.}
We append the activation maps generated by region-region matching to evaluate the performance of our model with region matching. As shown in Table \ref{mask}, we can conclude that the activation maps generated from region-to-region matching bring overall performance improvement. Compared with pixel-to-pixel matching, region-to-region matching can capture more contextual information, thus leading to improved performance.
\begin{table}[!h]
\centering
\caption{Ablation study on the different activation maps on the PASCAL-$5^i$ dataset.}
\resizebox{6cm}{!}{
\begin{tabular}{c|ccccc|c}
\toprule[2pt]
& \multicolumn{5}{c|}{1-shot mIoU} & \\ 
\multirow{-2}{*}{activation maps} & $5^0$ & $5^1$ & $5^2$ & $5^3$ & mean    
& \multirow{-2}{*}{FB-IoU} \\ \hline
block3      & 64.7   & 70.3   & 61.1 & 60.6   & 64.2  & 75.0 \\
block4      & 63.3   & 70.0   & 64.0  & 60.2   & 64.5  & 75.4  \\
block3\&4   & 64.5   & 70.1  & 64.3  & 60.4  & 64.8 & 75.9  \\
block3+region      &64.5    &69.4    & 64.0  & 60.6  & 64.6   &75.4  \\
block4+region      &62.9    &69.5    & 65.1   & 60.1  & 64.4  &75.2   \\
block3\&4+region   & 65.1 & 70.1 & 65.2 & 60.8 & 65.3 & 77.5   \\
\bottomrule[2pt]
\end{tabular}}
\label{mask}
\end{table}

\begin{figure*}[!h]
	\centering
	\includegraphics[width=\textwidth]{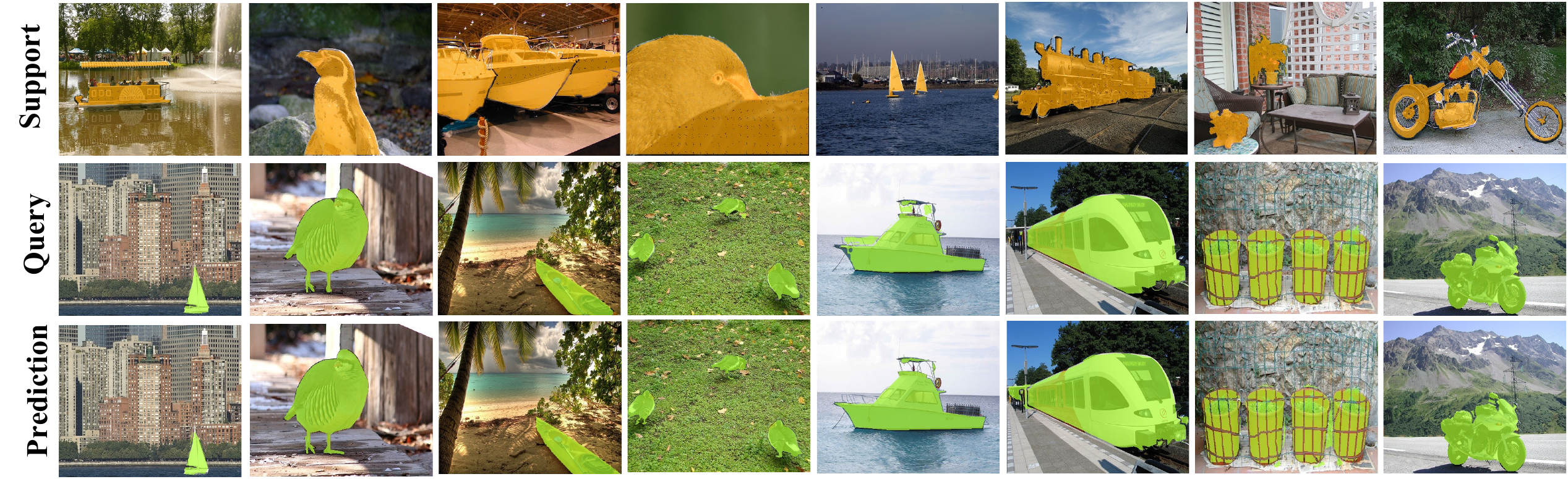}
	\caption{Qualitative results of the proposed SiGCN on PASCAL-$5^i$ benchmark with \textbf{large object appearance variations}. From top to bottom: Support Image \& Ground truth, Query Image \& Ground truth, and Query Prediction. Zoom in for details.}
	\label{pascal_app}
\end{figure*}

\begin{figure*}[!h]
	\centering
	\includegraphics[width=\textwidth]{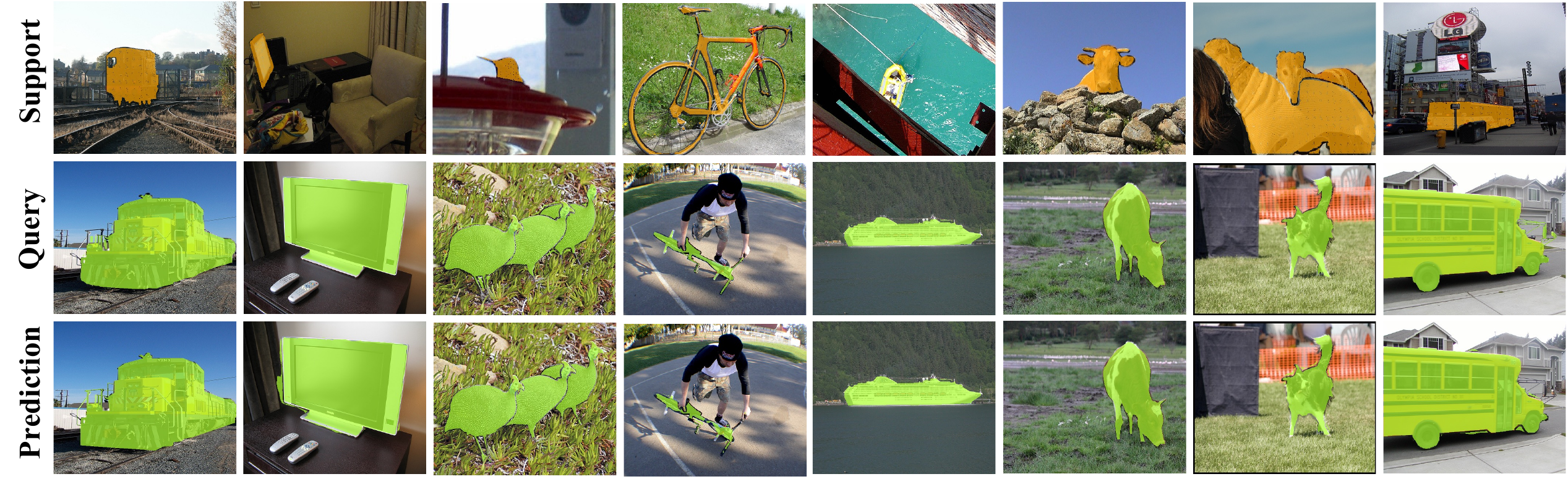}
	\caption{Qualitative results of the proposed SiGCN on PASCAL-$5^i$ benchmark with \textbf{large object scale variations}. From top to bottom: Support Image \& Ground truth, Query Image \& Ground truth, and Query Prediction. Zoom in for details.}
	\label{pascal_scale}
\end{figure*}

\begin{figure*}[!h]
	\centering
	\includegraphics[width=\textwidth]{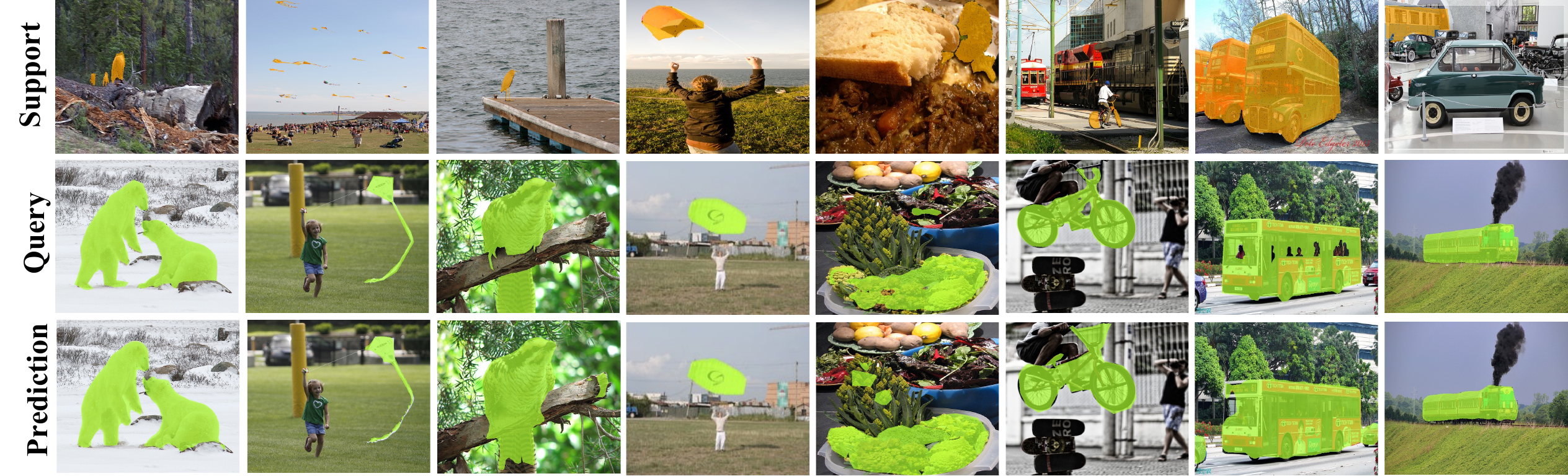}
	\caption{Qualitative results of the proposed SiGCN on COCO-$20^i$ benchmark with \textbf{large object appearance variations}. From top to bottom: Support Image \& Ground truth, Query Image \& Ground truth, and Query Prediction. Zoom in for details.}
	\label{coco_app}
\end{figure*}

\begin{figure*}[!h]
	\centering
	\includegraphics[width=\textwidth]{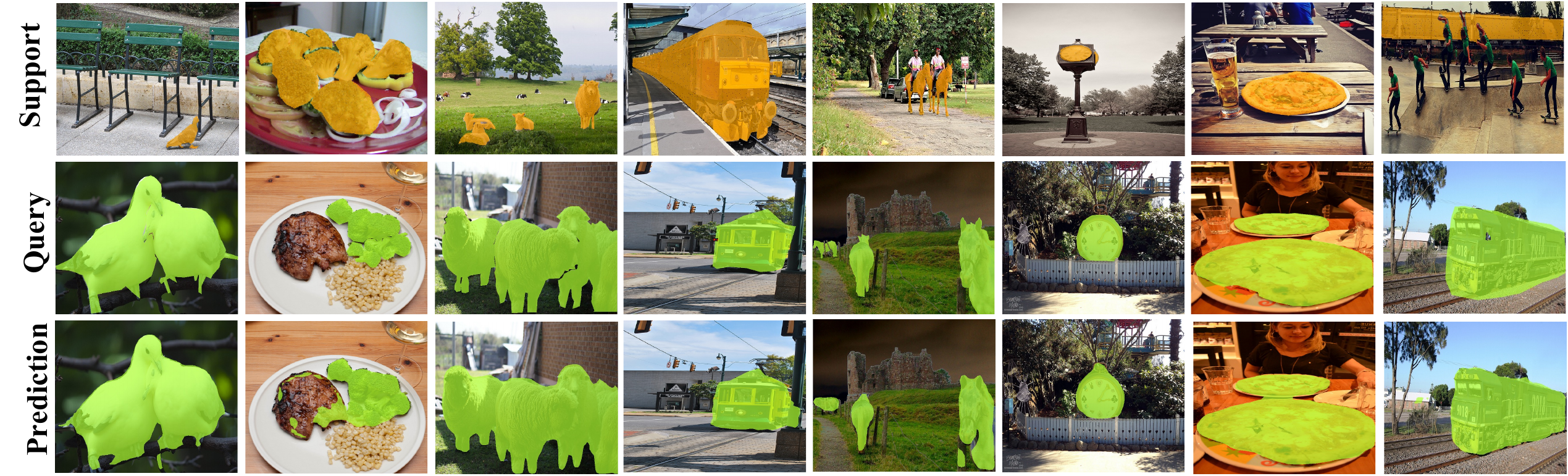}
	\caption{Qualitative results of the proposed SiGCN on COCO-$20^i$ benchmark with \textbf{large object scale variations}. From top to bottom: Support Image \& Ground truth, Query Image \& Ground truth, and Query Prediction. Zoom in for details.}
	\label{coco_scale}
\end{figure*}

\textbf{Additional Qualitative Results.}
We show more quantitative results of the proposed SiGCN in this section to further demonstrate its few-shot semantic segmentation performance. SiGCN aims to address the appearance variations problem in the few-shot semantic segmentation task, thus we show some examples sampled from PASCAL-$5^i$ and COCO-$20^i$ benchmarks with large appearance variations in Figure \ref{pascal_app} and Figure \ref{coco_app}, respectively. Furthermore, to demonstrate that our model can effectively address the scale variations problem, we give some examples from PASCAL-$5^i$ and COCO-$20^i$ benchmarks in Figure \ref{pascal_scale} and Figure \ref{coco_scale}, respectively. According to the quantitative results, we can conclude that our proposed SiGCN can effectively address the appearance and scale variations problem.

\end{document}